\documentclass{egpubl}
\usepackage{eurovis2025}

\SpecialIssuePaper         

\CGFccby

\usepackage{wrapfig}
\usepackage{color}
\usepackage{calc}
\usepackage{marginnote}
\usepackage{algorithm,algpseudocode}
\usepackage{amsmath}
\usepackage{titlesec}
\newlength\myheight
\newlength\mydepth
\settototalheight\myheight{Xygp}
\newcommand{\dbvisparagraph}[1]{\refstepcounter{paragraph}\noindent\textbf{#1\ --}\label{par:\theparagraph}}
\newcommand{\margingraphics}[1]{\marginnote{\includegraphics[angle=90,width=2em]{#1}}}

\definecolor{blue}{RGB}{72,22,105}
\definecolor{yellow}{RGB}{211,223,78}

\newcommand{\changed}[1]{\textcolor{black}{#1}}

\usepackage[T1]{fontenc}
\usepackage{dfadobe}

\biberVersion
\BibtexOrBiblatex
\usepackage[backend=biber,bibstyle=EG,citestyle=alphabetic,backref=true]{biblatex}
\electronicVersion
\PrintedOrElectronic
\ifpdf \usepackage[pdftex]{graphicx} \pdfcompresslevel=9
\else \usepackage[dvips]{graphicx} \fi

\titlespacing*{\section}    {0pt}{1.5ex plus 0.25ex minus .25ex}{-0.5ex plus .25ex}
\titlespacing*{\subsection} {0pt}{1.25ex plus 0.25ex minus .25ex}{-0.5ex plus .25ex}

\usepackage{egweblnk}

\title[LayerFlow]%
      {\textbf{\textit{LayerFlow}}: Layer-wise Exploration of LLM Embeddings\\ using Uncertainty-aware Interlinked Projections}

\author[
  Rita Sevastjanova,
  Robin Gerling, Thilo Spinner, and Mennatallah El-Assady
]
{\parbox{\textwidth}{\centering Rita Sevastjanova$^{1}$\orcid{0000-0002-2629-9579}, Robin Gerling$^{2}$, Thilo Spinner$^{1}$\orcid{0000-0002-1168-1804} 
        and Mennatallah El-Assady$^{1}$\orcid{0000-0001-8526-2613} 
        }
        \\
{\parbox{\textwidth}{\centering $^1$ETH Zurich, Switzerland\\
         $^2$University of Konstanz, Germany
       }
}
}

\begin{document}
\teaser{
  \centering

  \includegraphics[width=\linewidth]{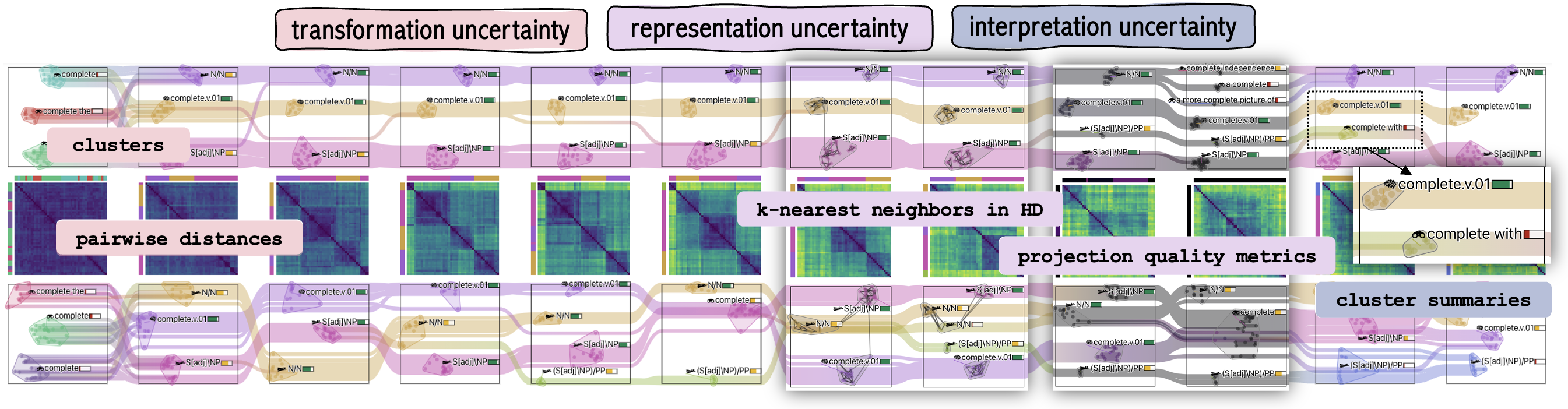}
  \vspace{-20pt}
  \caption{%
    \textbf{LayerFlow} supports the analysis of contextual word embedding properties. 
    To increase the awareness of the potential uncertainty within the transformation, representation, and interpretation steps of the used processing pipeline, we utilize multiple visual components such as cluster convex-hulls, pairwise distances, cluster summaries, projection quality metrics, and connections of k-nearest neighbors. 
  }
  \label{fig:overview}
}

\maketitle
\begin{abstract}
   Large language models (LLMs) represent words through contextual word embeddings encoding different language properties like semantics and syntax. Understanding these properties is crucial, especially for researchers investigating language model capabilities, employing embeddings for tasks related to text similarity, or evaluating the reasons behind token importance as measured through attribution methods. Applications for embedding exploration frequently involve dimensionality reduction techniques, which reduce high-dimensional vectors to two dimensions used as coordinates in a scatterplot. This data transformation step introduces uncertainty that can be propagated to the visual representation and influence users' interpretation of the data. To communicate such uncertainties, we present \textbf{\textit{LayerFlow}} -- a visual analytics workspace that displays embeddings in an interlinked projection design and communicates the transformation, representation, and interpretation uncertainty. In particular, to hint at potential data distortions and uncertainties, the workspace includes several visual components, such as convex hulls showing 2D and HD clusters, data point pairwise distances, cluster summaries, and projection quality metrics. 
   We show the usability of the presented workspace through replication and expert case studies that highlight the need to communicate uncertainty through multiple visual components and different data perspectives. 

\ccsdesc[500]{Human-centered computing~Visual analytics}
\ccsdesc[500]{Mathematics of computing~Dimensionality reduction}

\printccsdesc   
\end{abstract}

\section{Introduction}

In recent years, a large number of deep-learning-based language models (e.g., BERT~\cite{devlin2018}) have emerged, demonstrating remarkable performance in natural language processing (NLP) and understanding tasks. These models learn from large text datasets, acquiring language structures in an unsupervised manner. 
Thereby, they produce contextual word embeddings, representing words through vectors encoding different language properties. 
Extensive research has been conducted to understand the linguistic properties embedded in these vectors.
For instance, research indicates that BERT's middle layers capture syntactic features like dependency trees while early layers encode lexical features~\cite{rogers2020primer}. Analyzing these properties helps researchers better understand how language models process data and aids in developing models that generalize well, reducing biases and improving inclusivity.

A key focus in embedding analysis is visually exploring similarities to identify patterns in embedding clusters. The development of such visualizations has been an active field of study since the introduction of early static word embeddings like word2vec~\cite{mikolov2013distributed} and Glove~\cite{pennington2014glove}. Early visual methods focused on analogies~\cite{liu2017visual} and local word neighborhoods~\cite{heimerl2018interactive} and have since evolved to address properties of contextual language models including approaches that allow to inspect embedding neighborhoods~\cite{boggust2022embedding}, compare embeddings of multiple corpora~\cite{sivaraman2022emblaze}, models~\cite{sevastjanova2022adapters, boggust2022embedding}, or model's layers~\cite{sevastjanova2021,sevastjanova2022lmfingerprints}.
The current methods for embedding exploration often utilize dimensionality reduction techniques to reduce the high-dimensional (HD) vectors to two dimensions (2D) that are then used as coordinates to display words in a scatterplot.
Dimensionality reduction methods often exhibit well-known limitations. 
For instance, linear methods, like PCA~\cite{jolliffe2016principal}, are limited in detecting non-linear properties and can cause data points that are distant from each other in the HD space to appear close in the reduced space, potentially leading to a failure in preserving neighborhood relationships~\cite{MA201858}. Other methods, such as UMAP~\cite{mcinnes2018umap}, focus on preserving local neighborhoods but may fail to represent the distances between data points. Consequently, the resulting representations may not fully capture all true relationships and similarities within the HD data, introducing distortion. As emphasized by Nonato et al.~\cite{nonato2019multidimensional}, \textbf{possibility of distortions} between original and visual neighborhoods \textbf{introduces uncertainties} that impact the human analytic process. Haghighatkhah et al.~\cite{Haghighatkhah2022characterizing} in their work on text analysis separate this uncertainty into three types, i.e., \textit{transformation}, \textit{representation}, and \textit{interpretation} uncertainty. They show that due to \textit{transformation} steps like dimensionality reduction, the visual \textit{representation} might not be faithful to the underlying data, thus also influencing uncertainty in the analyst's \textit{interpretation} of the visualized word neighborhoods~\cite{Haghighatkhah2022characterizing}. 
Communication of potential uncertainties in the low-dimensional space is thus crucial for many use-cases involving dimensionality reduction, including word embedding analysis. 
According to Hullman et al.~\cite{hullman2019in}, accounting for uncertainty is critical to effectively reasoning about visualized data.
Various ways of visualizing such uncertainties have been explored in prior work~\cite{padilla2020uncertainty,amit2020uncertainty,fernandes2018uncertainty}, whereby much focus has been put into communicating the uncertainty in projection visualizations (see, e.g.,~\cite{amit2020uncertainty,nonato2019multidimensional,aupetit2007visualizing}).

In this paper, we aim to communicate the uncertainty in the embedding exploration process, addressing the uncertainty introduced by transformation, representation, and interpretation steps~\cite{Haghighatkhah2022characterizing}.
Building upon our prior work that introduces an interlinked projection approach for embedding investigation~\cite{sevastjanova2021}, we present \textbf{LayerFlow}, an embedding exploration workspace that combines multiple visual components for uncertainty communication.
In this work, we support the combination of 2D and HD clusters, pairwise distance representations, and cluster summaries, as well as incorporate a set of projection quality metrics and representations of k-nearest neighborhoods to increase the user awareness of potential projection distortions.

To summarize, we present (1) \textbf{LayerFlow}, 
a novel workspace for embedding exploration utilizing multiple visual components for communication of data transformation, representation, and interpretation uncertainty;
(2) evaluation through case studies and expert interviews, along with a replication of findings from previous research on the linguistic properties encoded in embeddings.

\section{Related Work}
In the following, we describe the prior work related to embedding visualization and projection uncertainty.

\subsection{Word Embedding Visualizations}
Since the development of the first static word embedding models (e.g., word2vec \cite{mikolov2013distributed} and Glove \cite{pennington2014glove}), visualization methods have been developed to support the analysis of embedding properties. The first tools facilitated analogies \cite{liu2017visual} and the analysis of word local neighborhoods \cite{heimerl2018interactive}.
The development of contextual language models introduced new aspects that were incorporated into visualizations, e.g., the properties of the word contexts.
Berger~\cite{berger2020visually} explored correlations between embedding clusters in BERT~\cite{devlin2018}.
EmbComb~\cite{heimerl2020embcomp} employs diverse metrics to measure variations in the local structure around embedding objects, such as tokens. 
Embedding Comparator~\cite{boggust2022embedding} enables comparing embeddings through small multiples, while calculating and visualizing similarity scores based on shared nearest neighbors in their local neighborhoods. 
Emblaze~\cite{sivaraman2022emblaze} utilizes an animated scatterplot and incorporates visual augmentations to represent changes in the analyzed embedding spaces.
Our prior work, {LMFingerprints}~\cite{sevastjanova2022lmfingerprints}, applies scoring techniques to examine properties encoded in embedding vectors, while another approach~\cite{sevastjanova2022adapters} proposes comparative visualizations to explore embedding encoded biases.
The most related work to this paper is our LMExplorer~\cite{sevastjanova2021} interface that utilizes interlinked projection visualization for analyzing word embedding self-similarity.
This paper presents an approach that builds on this interlinked projection visualization and proposes multiple visual components for uncertainty communication~\cite{Haghighatkhah2022characterizing}.

\begin{figure}[t]
  \centering
  \includegraphics[width=\linewidth]{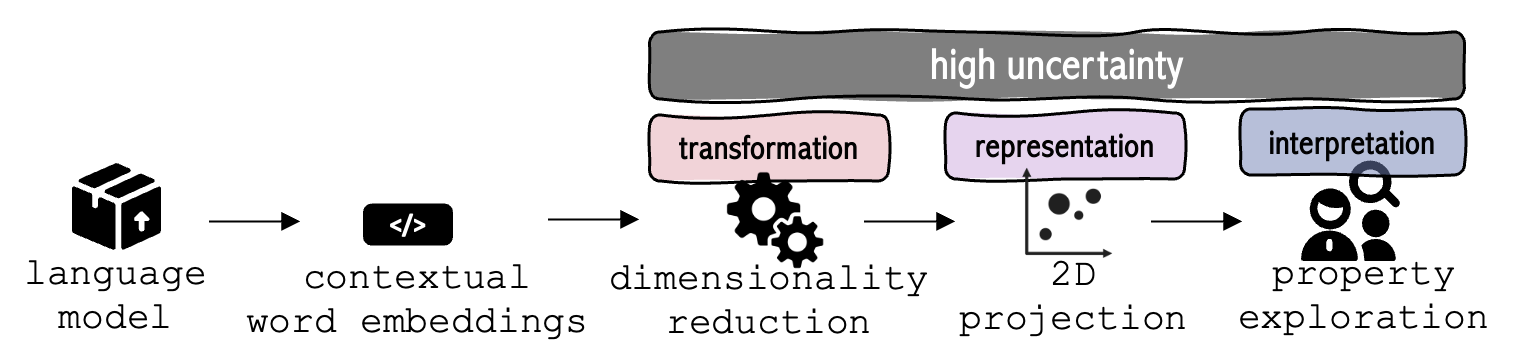}
  \vspace{-18pt}
  \caption{%
    Many approaches for word embedding analysis utilize a 2D projection of the HD vectors to be visualized for further investigation. These processing steps can introduce uncertainties related to the transformation, representation, and interpretation steps.
  }
  \label{fig:uncertainty}
  \vspace{-2.2em}
\end{figure}

\subsection{Visualizing Projection Uncertainty}
Visualizing uncertainty is a widely researched topic \changed{as shown by Amit et al. \cite{amit2020uncertainty} in their interactive survey.} Diverse methods such as explicit displays, animations, overlay visualizations, summary statistics using visual variables, and uncertainty as an additional data dimension can be used to communicate the uncertainty to the users \cite{padilla2020uncertainty}. 
\changed{Communicating uncertainty is important} for embedding exploration tools that often utilize projection methods to represent the HD vectors in scatterplot visualizations.
The dimensionality reduction methods are prone to artifacts; the reduction typically involves some loss of information that is mimicked in the scatterplot and ill-represented data point pairwise distances.
Thus, diverse visualization methods have been designed to communicate such uncertainties in the projected data \cite{aupetit2007visualizing}.
Additional visualizations, such as \changed{matrices}, can be used to display the differences in the distances between the data points in the HD and 2D space \cite{cutura2020comparing}.
Nonato et al. \cite{nonato2019multidimensional} describe different layout enrichment methods to communicate uncertainty in projections. 
These methods utilize the closeness of similar instances in the visual space to convey additional information related to specific instances or groups of instances. The authors distinguish between direct, cluster-based, and spatially structured enrichment methods.
Another group of methods includes layout enrichment for distortion analysis, for instance, by visualizing trustworthy and unreliable regions in the projection.
Several approaches exist that utilize layout enrichment for content analysis, including UTOPIAN \cite{choo2013utopian} and FaceAtlas \cite{Cao2010FacetAtlasMV} as well as enrichment for distortion analysis (e.g., Probing Projections \cite{stahnke2016probing} and ProxiViz \cite{aupetit2007visualizing}). \changed{In this paper, we aim to highlight the uncertainty in the produced 2D neighborhoods and thus first unfold the 2D space through a Sankey diagram design, creating additional space to integrate visual overlay elements. 
Second, we integrate these overlay elements, e.g., in the form of visual connections of k-nearest neighbors, color-encoded projection quality metrics, or cluster-based summaries of the embedding properties.}

\begin{figure*}[th]
  \centering
  \includegraphics[width=\textwidth]{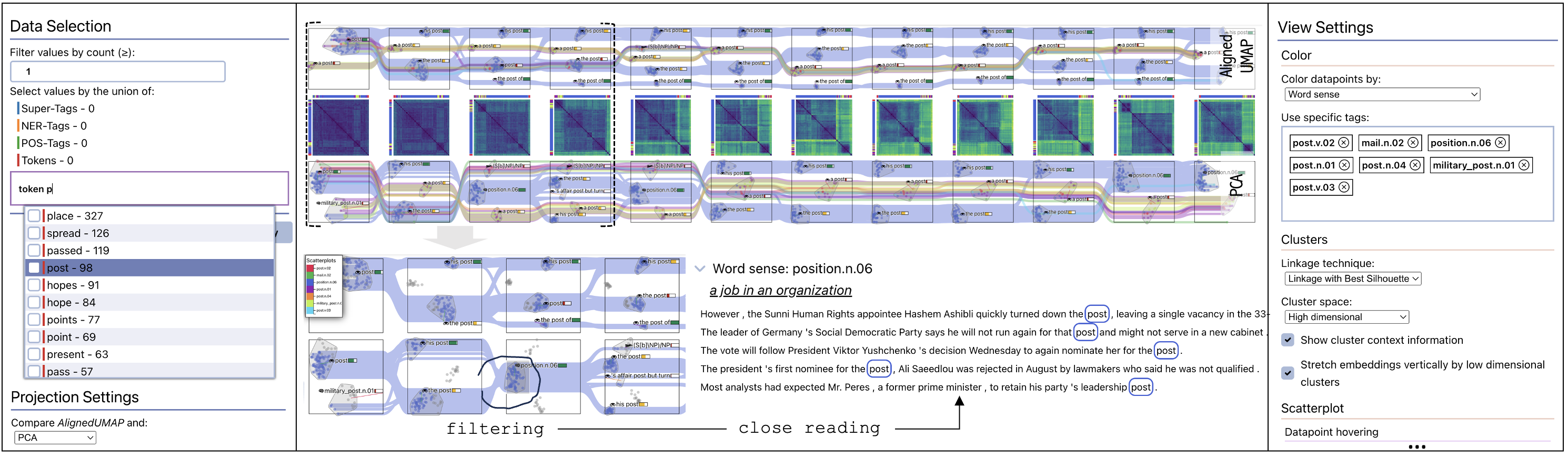}
  \vspace{-20pt}
  \caption{%
    The \textbf{LayerFlow} workspace consists of two setting sidebars and one main view. In the settings, the user can select the data, specify the projection techniques and properties for the data visualization (e.g., color encoding, clustering methods, and further parameters). The interlinked projections for one or two projection algorithms (e.g., PCA, UMAP, Aligned UMAP) are displayed in the middle of the view. The interactively selected data points are highlighted and further displayed in a close reading view for in-detail inspection.
  }
  \label{fig:workspace}
  \vspace{-2.2em}
\end{figure*}

\section{Problem Characterization}
In the following, we describe the target users, their tasks when analyzing word embedding vectors, \changed{and requirements for a visual analytics workspace. 
The tasks and requirements were defined based on feedback from our long-term collaboration with three NLP experts, derived from their interactions with our prior interfaces \cite{sevastjanova-etal-2021-explaining, sevastjanova2022lmfingerprints}, and insights from a literature review.}

\subsection{Users and Tasks}
Our objective is to facilitate developers and researchers in their efforts to understand word embedding properties of deep-learning-based language models. 
\changed{
Based on our observations from collaborating with NLP researchers over the past years, the researchers typically aim to (\textbf{T1}) \textit{analyze embeddings based on their similarity and common characteristics}; (\textbf{T2}) \textit{analyze property changes between layers to investigate how the information captured in embedding weights propagates through the model's architecture}; and (\textbf{T3}) \textit{assess the reliability of visualization patterns to determine when they can be trusted for drawing conclusions about model properties.}
}
The \textbf{T3} is critical since as depicted in \autoref{fig:uncertainty}, many approaches related to contextual word embedding analysis include the generation of a 2D projection of the HD vectors.
The process of mapping HD data to 2D coordinates can introduce distortions and uncertainties, as emphasized by a wide range of prior work (see, e.g., \cite{nonato2019multidimensional,amit2020uncertainty,padilla2020uncertainty}).
Similar challenges have been described by Haghighatkhah et al. \cite{Haghighatkhah2022characterizing} when it comes to text analysis approaches and uncertainties in mapping HD text data (e.g., word embeddings) to 2D coordinates. In addition to other uncertainties related to text semantics, these methods include uncertainties that occur during the data transformation and visualization steps, \changed{i.e., the \textit{transformation}, \textit{representation}, and \textit{interpretation} uncertainty.}

\subsection{Requirement Analysis}
We describe the three main requirements (Rs) for an uncertainty-aware embedding exploration workspace.

\noindent{\textbf{Transformation Uncertainty --}}\label{sec:transformation}
When analyzing contextual word embeddings, the data needs to be prepared for visualization purposes. Although some methods exist that inspect the properties of single weights in the embedding vectors (see the survey on neuron analysis~\cite{Sajjad2022survey}), which does not require data transformation, most of the current approaches focus on either embedding similarity analysis in the HD space (e.g., through similarity measures \cite{ethayarajh2019contextual,sevastjanova2022lmfingerprints} or clustering \cite{berger2020visually}) or embedding reduction to 2D and visual exploration (e.g., \cite{boggust2022embedding,sevastjanova2022adapters}).
This data transformation step is typically performed through a dimensionality reduction method, which comes with uncertainties, e.g., some methods preserve distances (e.g., MDS) while others -- local and/or global neighborhoods (e.g., t-SNE, UMAP). Thus, uncertainty can occur with regard to the distance or topology preservation.
Nonato et al. \cite{nonato2019multidimensional} write that it is exceedingly challenging to represent the data as a set of points in the visual space while perfectly preserving euclidean distances.
\textit{To communicate the transformation uncertainty, we should represent the data point pairwise distances in both 2D and HD space} \textbf{(R1)}.

\noindent{\textbf{Representation Uncertainty --}}
The created projection space is one of numerous potential representations of the original HD data. Aupetit et al. \cite{aupetit2007visualizing} write that ``The projection is neither true nor false, but it may be faithful or not according to the original data w.r.t. to some geometrical or topological criterion. The unknown degree of faithfulness of the projection is then a source of uncertainty in the visualization.''
According to Nonato et al. \cite{nonato2019multidimensional}, scatterplots representing projected data typically contain uncertainties caused by clutter, resolution, and contrast.
Furthermore, uncertainty can arise due to the pre-attentive perception of the relative position of the points.
If data insights on word embedding properties are made using such a visual projection space
, it is crucial to communicate this uncertainty, e.g., through quality metrics. 
\textit{To communicate the representation uncertainty, we should highlight potential distortions in the visualized projection space} \textbf{(R2)}.

\noindent{\textbf{Interpretation Uncertainty --}}
The patterns depicted in the scatterplot may not be sufficient for a comprehensive understanding of the HD data, requiring the advancement of data representation \cite{nonato2019multidimensional}.
To support users in interpreting data and reducing uncertainty, we can use layout enrichment strategies  \cite{nonato2019multidimensional}.
Also, in the context of word embedding analysis, the interpretation of the produced data layouts and clusters can be challenging since multiple factors can influence the embedding similarity. The BERTology paper \cite{rogers2020primer} and our prior work \cite{sevastjanova2022lmfingerprints} list a range of properties that can be encoded in embedding vectors and thus influence their similarity.
These lists are not exhaustive as language models theoretically have the potential to capture additional aspects that humans might not anticipate, given the disparity between our mental models and the language representation within these models \cite{Sevastjanova2022BewareTR}.
Thus, it is challenging to design representations that guarantee a high level of certainty in interpretation. Nevertheless, hints to possible reasons can reduce the degree of such uncertainty.
\textit{To address the interpretation uncertainty, we should include layout enrichment methods that provide additional, semantic information on possible reasons for the displayed visual patterns} \textbf{(R3)}.

\section{LayerFlow Workspace: Data Modeling}

In the following, we describe the used data pre-processing and modeling steps to prepare the data for visualization.

\subsection{Data Pre-Processing}\label{sec:pre-processing}
We extract contextual word embedding vectors layer by layer for every token-context pair in the corpus. We define context as a sequence of tokens encompassing a single sentence.

\noindent\textbf{Dimensionality Reduction --} PCA, UMAP, and Aligned UMAP \cite{aligned_umap} algorithms are used to reduce the HD vectors to 2D coordinates. We integrate them into the workspace to explore the differences between a linear (PCA) and non-linear (UMAP) methods and their difference to an aligned algorithm (Aligned UMAP), which is particularly representative for our use case of computing projections on a temporarily evolving data (i.e., layer-wise embeddings).

\noindent\textbf{Data Point Annotation --}
Each data point is annotated with multiple features used for further processing steps. 
We extract Part-of-Speech (POS) tags using the NLTK library \cite{bird-loper-2004-nltk}, and syntactic categories using a Combinatory Categorial Grammar library \cite{ccgjs}. 
We use GlossBERT \cite{glossbert} 
model for word sense disambiguation to assign each word its unique word sense based on its context and spaCy \cite{spacy_ner} 
to extract the named entity categories.

\subsection{Data Modeling for Uncertainty Communication}\label{sec:data-modeling}

To effectively communicate the uncertainty in the different steps of the analysis process, we need to prepare the input data for visualization. This includes clustering methods, distance, quality metric, and k-nearest neighbor computation, and cluster annotation.

\margingraphics{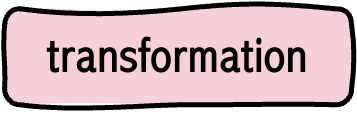}
\dbvisparagraph{Clustering in 2D and HD Space}~\changed{After the data is reduced to two dimensions using a dimensionality reduction method, we apply a hierarchical clustering approach utilizing cosine distance to determine clusters in the 2D as well as HD space  (supports~\textbf{R1} in~\autoref{sec:transformation})}. 
Our approach supports various hierarchical linkage methods, including single, complete, average, and weighted linkage. To determine the optimal number of clusters, we identify the cut in the dendrogram that achieves the highest silhouette score among the top 10\% of all possible cuts (i.e., the upper 10\% cuts in the dendrogram).
Instead of using the same linkage strategy for all projections (i.e., all model layers), we select the linkage technique with the best silhouette score for each layer separately.

\dbvisparagraph{Distances in 2D and HD Space} To communicate the distortions and differences between the original HD data and the transformed 2D data (\textbf{R1}), we compute and store the pairwise \changed{cosine} distances between single data points in the 2D and HD space, respectively.

\margingraphics{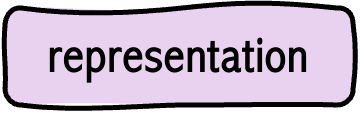}
\dbvisparagraph{Projection Quality Metrics} To show the representation uncertainty (\textbf{R2}), we compute projection quality metrics, 
i.e., \changed{\textit{Projection Precision Score} \cite{schreck2010techniques}, \textit{Point Compression} \cite{aupetit2007visualizing}, \textit{Point Stretching} \cite{aupetit2007visualizing}, \textit{Aggregated Projection Error} \cite{aupetit2007visualizing}, \textit{True Neighbors} (\textit{Neighborhood Preservation}) \cite{vernier2020quantitative}, \textit{False Neighbors} \cite{martins2014visual}, \textit{Missing Neighbors} \cite{martins2014visual} (aka. \textit{Trustworthiness} \cite{vernier2020quantitative}), and \textit{Local Continuity Meta-Criterion} \cite{chen2009local}.}

\changed{The first four metrics are computed based on pairwise distances between data points.
This becomes challenging when applied to PCA or UMAP variants, as neither method guarantees the preservation of such pairwise distances.
Thus, even if the 2D clusters visually align with those in the HD space, their quality may still be low due to inaccurately represented distances.
Other metrics rely on the \textit{k} nearest neighbors either with a fixed parameter \textit{k} for the whole projection \cite{vernier2020quantitative, martins2014visual}, or, alternatively, different \textit{k} values for each data point based on the HD cluster size in which it is contained \cite{chen2009local}.
In both cases, the metrics utilize euclidean distance to compute the nearest neighbors, which would produce low scores for data points that are located near other clusters. 
} 
\changed{To overcome these limitations, we introduce two additional metrics, i.e., 
\textit{False Positive Rate} (FPR) and \textit{False Negative Rate} (FNR) that neither use a specific \textit{k}, nor the euclidean radius.
Instead, they measure the projection quality by taking the minimum spanning tree (MST) of the 2D space into account. In particular, we represent the data points in 2D as a fully connected graph where edges represent the distances between the data points. We compute the MST on this graph by applying the Kruskal’s algorithm \cite{kruskal1956shortest}. 
As showcased in \autoref{algorithm},
for each data point (target), the algorithm measures the ratio of the target's \textit{n} closest points in the MST (\textit{n} stands for the size of the target's HD cluster) that do not exist in the target's HD cluster (FPR) and the ratio of data points in the target's HD cluster that are not among the target's closest points in the MST (FNR). 
} 

\begin{algorithm}[t]
\caption{FPR and FNR Quality Metrics}
\label{algorithm}
\begin{algorithmic}[1]
\State \textbf{Input}: HD clusters, MST of a fully connected graph in the LD space (data points as nodes, pairwise distances as edges)
\For{each point $p$ in the LD space}
    \State $n \gets |C(p)|$ \Comment{Size of HD cluster}
    \State $sp \gets p$ \Comment{Start point}
    \State $NN(p) \gets \emptyset$ \Comment{Initialize nearest neighbors set}
    \For{$i \gets 1$ to $n-1$}
        \State $nn_i \gets \arg\min_{q \in \text{MST} \setminus NN(p)} d(NN(p) \cup \{sp\}, q)$
        \State $NN(p) \gets NN(p) \cup \{nn_i\}$
    \EndFor
\EndFor
\State Compute FPR, and FNR as for confusion matrix, i.e.:
\State $TP = |\{q : q \in C(p) \cap NN(p)\}|$ \Comment{True Positives}
\State $TN = |\{q : q \not\in C(p) \cap \overline{NN(p)}\}|$ \Comment{True Negatives}
\State $FP = |\{q : q \not\in C(p) \cap NN(p)\}|$ \Comment{False Positives}
\State $FN = |\{q : q \in C(p) \cap \overline{NN(p)}\}|$ \Comment{False Negatives}
\State 
$FPR = \frac{FP}{FP + TN}$, $FNR = \frac{FN}{FN + TP}$ 
\end{algorithmic}
\end{algorithm}

\dbvisparagraph{K-Nearest Neighbors in HD Space} To visually mark the differences in nearest neighborhoods in the 2D and HD space (\textbf{R2}), we compute and store the top-k nearest neighbors in the HD space by using the cosine similarity function on the HD vectors.

\margingraphics{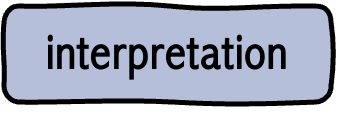}
\dbvisparagraph{Cluster Summaries in 2D and HD Space} To reduce the uncertainty that is typically caused by the user’s subjective interpretation of the visualization (\textbf{R3}), we compute summaries of the most common features for the data points within each 2D and HD cluster.
As shown by our prior work \cite{sevastjanova2022lmfingerprints}, different properties of embedding neighborhoods can be measured, such as lexical, surface, and context features. In this work, we design summaries based on the common POS tags, syntactic categories, n-grams, \changed{named entities, and token indices}. 
The most common feature in the token contexts (two preceding and two following words) within a cluster is determined and used as the cluster's summary. We further compute the certainty of the particular summary. \changed{It shows whether the summary label (e.g., NOUN) is representative for all data points in the cluster and whether it is unique for this very specific cluster, and is computed as follows:}

\vspace{-5pt}
\begin{math}
\text{Certainty} = \left( \frac{\# \,  {\text{label x in cluster}}}{\# \, {\text{label x in projected data}}} \right)^2 \cdot \left( \frac{\# \, {\text{label x in cluster}}}{\text{cluster size}} \right)^2
\end{math}
\vspace{-5pt}

\section{LayerFlow Workspace: Design Rationale}\label{sec:design-rationale}
The workspace consists of two setting sidebars, and one main view, which is displayed in the middle of the screen.
The users can first specify the words for exploration through the data selection menu and select the projection techniques with their parameters (e.g., $n\_neighbors$, $min\_dist$ for UMAP). 
The words can be filtered based on strings or their annotations, e.g., POS tags.
After the projections are computed and visualized (see \autoref{sec:interlinked-projections}), the visual components for uncertainty communication (see \autoref{sec:visual-components}) can be selected in the sidebar on the right-hand side of the screen.

\subsection{Interlinked Projections}\label{sec:interlinked-projections}
The main visual component of the workspace is interlinked projections, which was introduced in our prior work \cite{sevastjanova2021} for word embedding investigation tasks.
In this visualization, the individual layers are visualized as scatterplots, one scatterplot representing embeddings extracted from one layer.
The data points are linked between the layers to enable tracing their positional
changes.
The workspace supports various interaction methods.
For instance, the brushed data points in the projection are filtered and highlighted in all visual representations among all layers.
By hovering over a datapoint, the word's context is displayed in a tooltip.
\changed{To enable projection comparison}, the user can display two projection techniques simultaneously, as shown in \autoref{fig:workspace}.

\begin{figure}[t!]
  \centering
  \includegraphics[width=\linewidth]{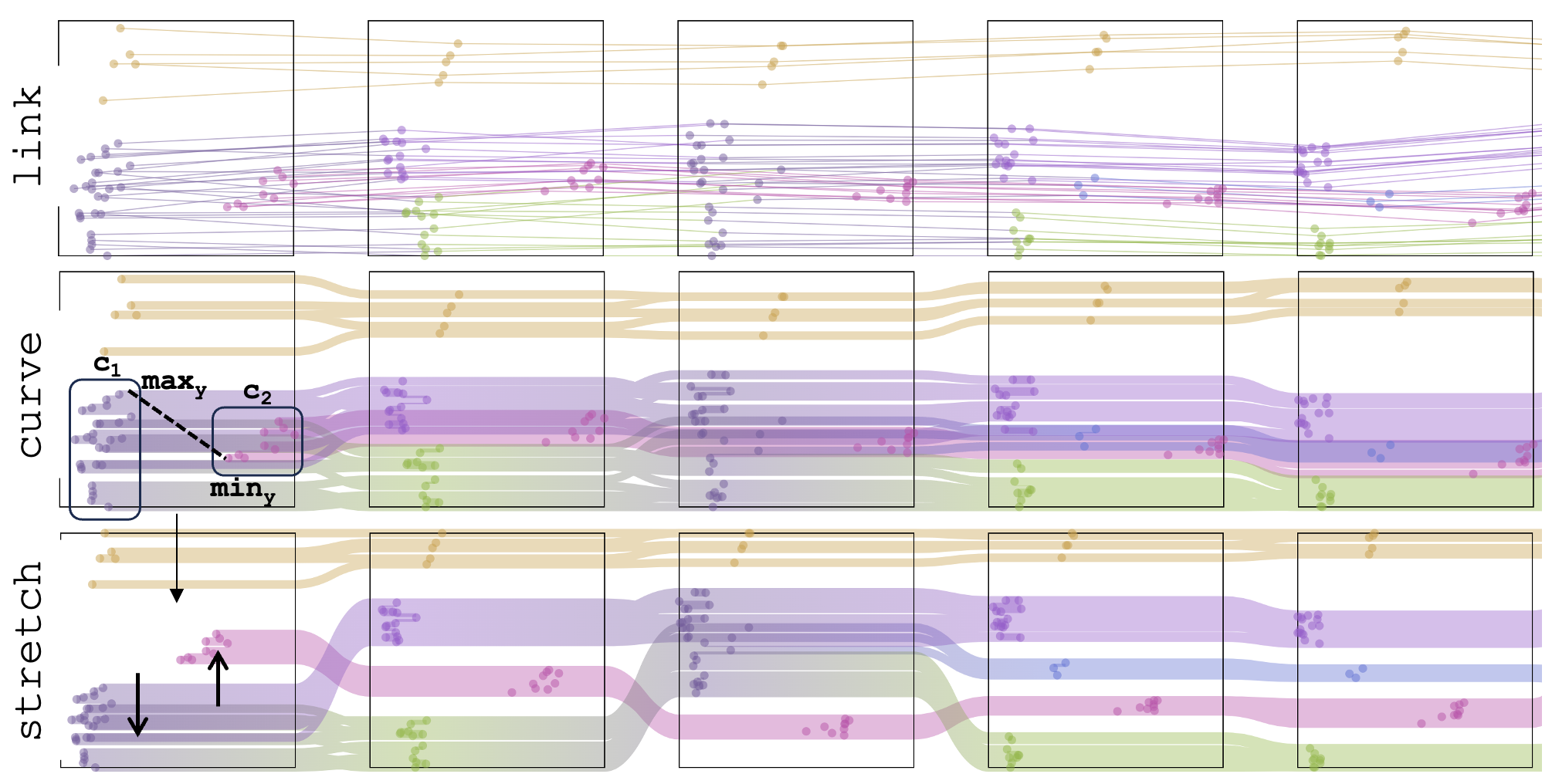}
  \vspace{-18pt}
  \caption{%
    To increase the readability of the word positional changes between consecutive layers, we use a Sankey layout that links the data points of the detected 2D clusters. 
  }
  \label{fig:flow-vis-design}
  \vspace{-2.5em}
\end{figure}

The original design of the interlinked projections \cite{sevastjanova2021} utilizes an edge bundling method to group lines of the same clusters, reducing the visual clutter. Edge bundling comes with limitations, e.g., the clusters can result in ambiguous connections that do not exist in the data \cite{Wallinger2021EdgePathBA}. In this work, we adapt the visual design to improve the investigation of neighborhood and cluster changes between the consecutive layers by utilizing a Sankey layout. 
Sankey diagrams have been used in the prior work for different use cases, including the analysis of neural networks. For instance, Halnaut et al. \cite{halnaut2020deep} use a flow chart to trace the progressive classification of a neural network, while Puhringer et al. \cite{puhringer2020instanceflow} use it to visualize the flow of instances throughout epochs. 
We utilize the idea of the Sankey diagram and define the flows by the cluster positions in the 2D space. The algorithm is implemented in three main steps, as sketched in \autoref{fig:flow-vis-design}.

\noindent{\textbf{Step 1: Interlinked Scatterplots --}} The initial phase mirrors the approach outlined in our prior work \cite{sevastjanova2021}. The projections are arranged horizontally adjacent to each other, with a gap of 50 pixels between the scatterplots. To track the positional changes of data points across successive layers, we establish connections between a data point across all layers using polylines.
Data points and polylines can be colored according to the different properties introduced in \autoref{sec:pre-processing} (e.g., POS tags or word senses).

\noindent{\textbf{Step 2: \changed{Sankey Diagram --}}} 
To increase the readability of the data point positional changes, 
we increase the line thickness and \setlength{\columnsep}{5pt}%
\begin{wrapfigure}[5]{r}{0.25\textwidth}
  \centering
  \vspace{-12pt}
  \includegraphics[width=\linewidth]{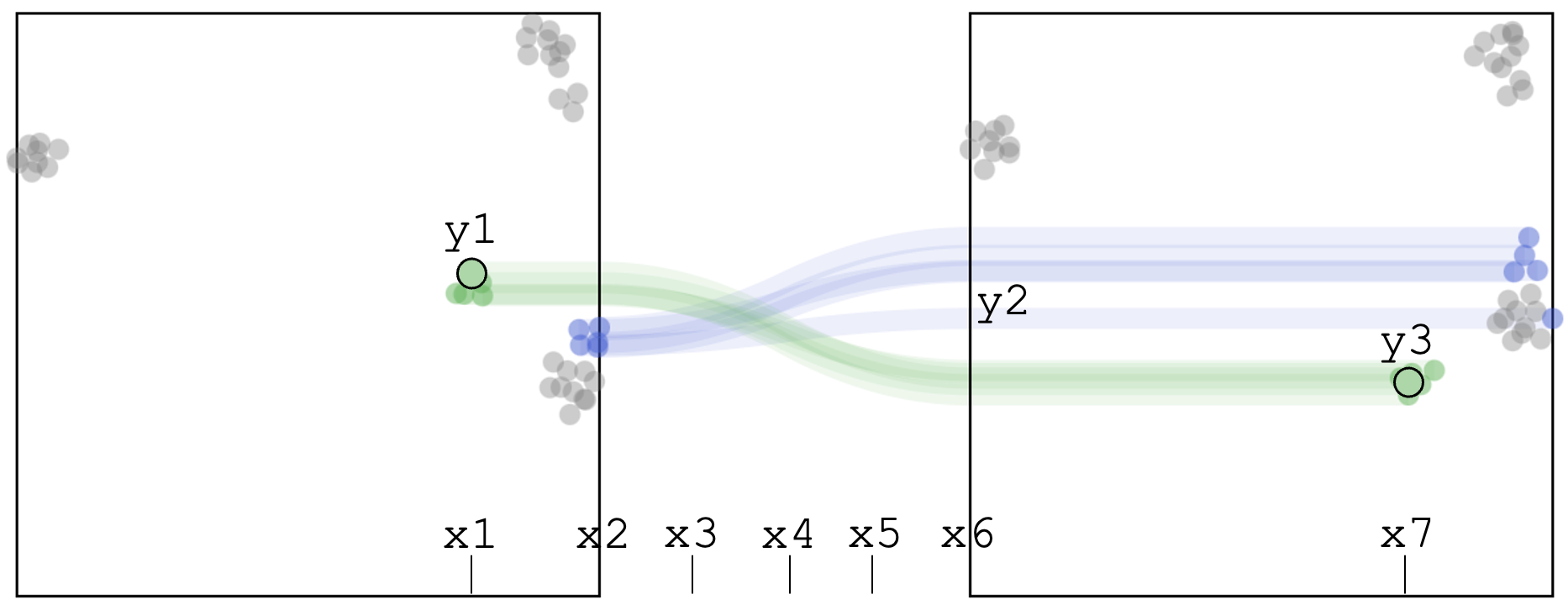}
\end{wrapfigure}reduce their opacity.
Moreover, instead of displaying straight lines between two data points, we create six path elements.
The first path element is a horizontal line from the source data point to the layer's right border; see $(x1, y1) \rightarrow (x2, y1)$ in the side figure.
The second path element is a cubic bézier curve from the border to the horizontal center between the two layers with the vertical center between the data points y-value in the
previous and the next layer $(x2, y1) \rightarrow (x4, y2)$ with a control
point at $(x3, y1)$. 
It is followed by a horizontally and vertically mirrored cubic bézier $(x4, y2) \rightarrow (x6, y3)$ with a control point at $(x5, y3)$. 
The fourth path element is created as in the first step but from the left border of the next layer's scatterplot to the data point in the next layer $(x6, y3) \rightarrow (x7, y3)$.

In this representation, the lines of data points that are located in close proximity overlap and thus produce visual artifacts such as \setlength{\columnsep}{5pt}%
\begin{wrapfigure}[5]{r}{0.25\textwidth}
  \centering
  \vspace{-12pt}
  \includegraphics[width=\linewidth]{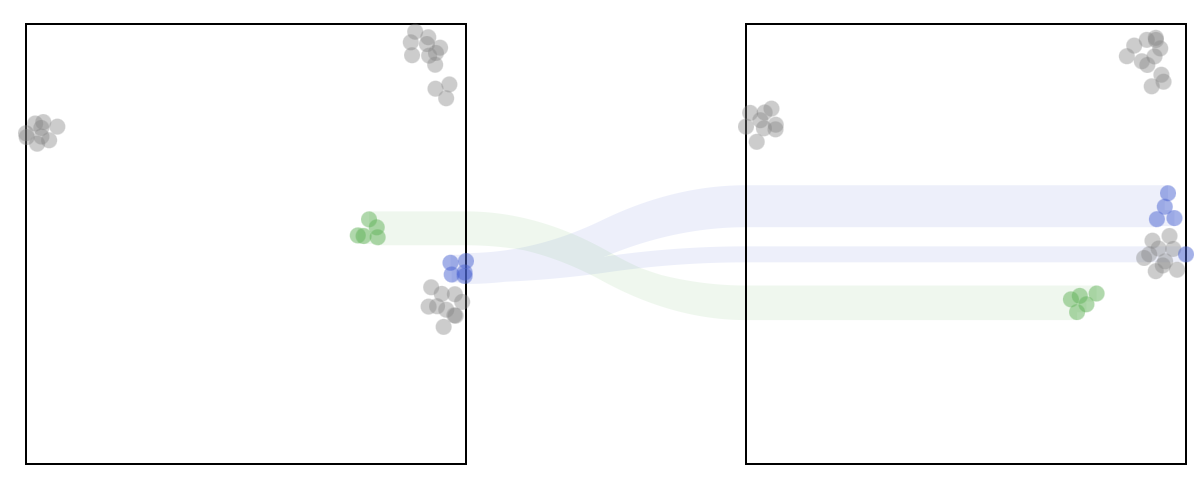}
\end{wrapfigure}darker color regions. In order to reduce such visual artifacts, we create one path element for data points that share the neighborhood and data properties (e.g., POS tag) in two consecutive layers using the d3.js path generator function that concatenates the individual paths into one string. As shown in the side figure, this produces a less cluttered representation whereby data points that change their cluster or have different properties than their neighborhoods become more visible. 

\noindent{\textbf{Step 3: Flow Stretching --}} The visualization produced in step 2 has a limitation. As shown in \autoref{fig:flow-vis-design} ({curve}), if clusters in the projection are placed horizontally next to each other, the curved lines of these clusters will overlap, limiting their readability.
Thus, we introduce a vertical cluster stretching. As shown in \autoref{fig:flow-vis-design} ({stretch}), 2D clusters that share their \textit{y} coordinates within a scatterplot are moved vertically until the overlap is reduced. First, the clusters are sorted by median. Then, for each two successive clusters $c_1$ and $c_2$ in the sorted array, we compare $max_y(c_1)$ with $min_y(c_2)$ and move $c_2$ upward if $\min_y(c_{2}) \leq \max_y(c_1)$. 
\changed{The pseudocode is given in the supplementary material and a visual explanation in \autoref{fig:flow-vis-design}.}
\textit{Remark}: This method distorts the global distances in the original projection. Nevertheless, in this work, we aim to highlight the uncertainty in the produced 2D neighborhoods while reducing the visual clutter, which is guaranteed by this method.  

\subsection{Visual Components for Uncertainty Communication}\label{sec:visual-components}
We use the output of the data modeling step (\autoref{sec:data-modeling}) to design the visual components for the communication of uncertainty.

\begin{figure}[b!]
  \centering
  \vspace{-12pt}
  \includegraphics[width=\linewidth]{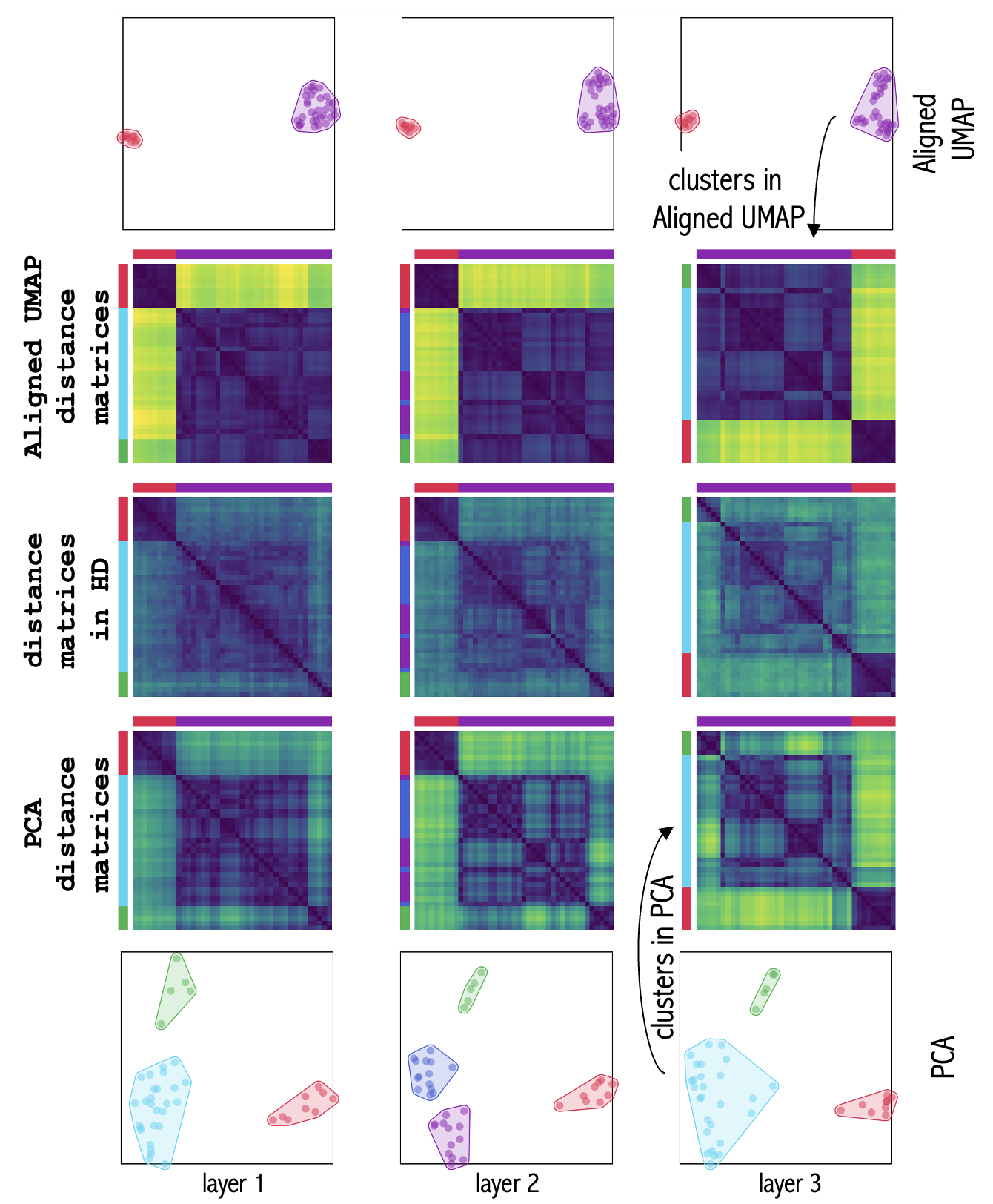}
  \vspace{-18pt}
  \caption{%
    Distance matrices can be used to communicate the transformation uncertainty. Users can observe clusters visible in the HD space but not separated in the 2D space and vice versa. 
  }
  \label{fig:distance-matrices}
\end{figure}

\begin{figure*}[t]
  \centering
  \includegraphics[width=\linewidth]{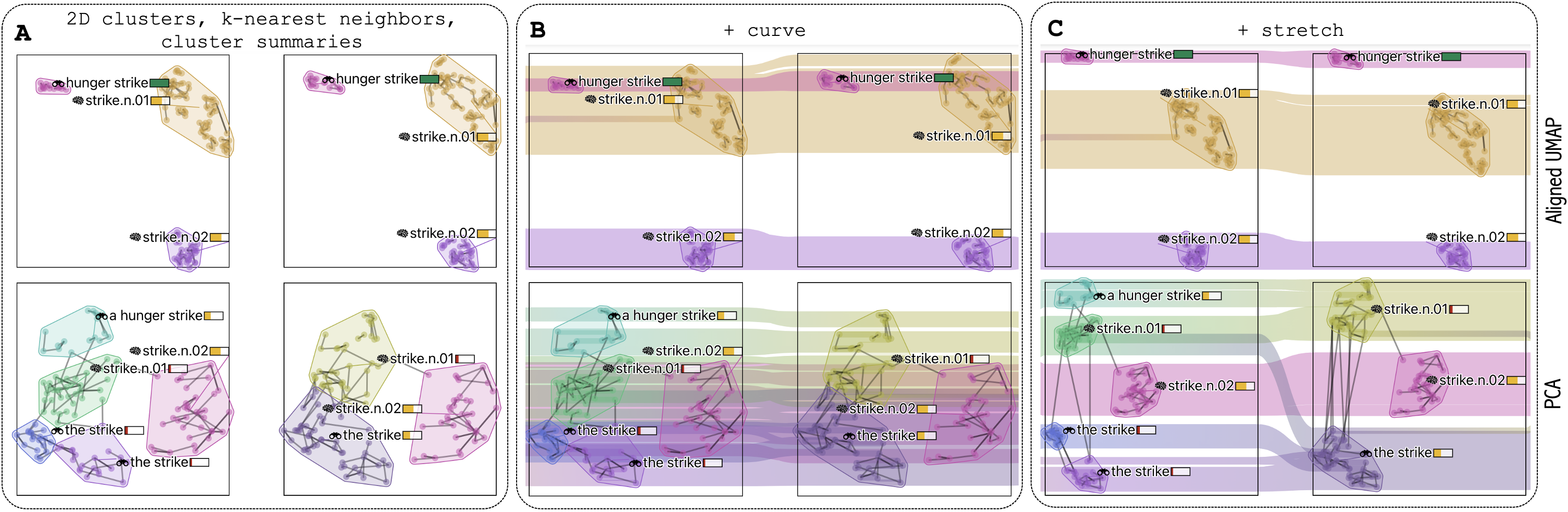}
  \vspace{-18pt}
  \caption{%
    A combination of visual components for uncertainty communication. The representation uncertainty is shown in \textbf{(A)} through the connection of lines of k-nearest neighbors (here, k=1) in the HD space. 
    In addition, cluster summaries highlight whether the data points in the 2D clusters have some linguistic concepts in common. Here, Aligned UMAP creates a cluster of tokens related to \texttt{hunger strike} with high certainty (see the green bar next to the label). The visual design of \textbf{LayerFlow} helps to trace token position changes within the consecutive layers. By bundling \textbf{(B)} and stretching \textbf{(C)} the flows, it becomes apparent that PCA faces challenges in generating distinct clusters that align with HD neighborhoods, and furthermore, there are more frequent cluster transitions compared to the output of the Aligned UMAP.
  }
  \label{fig:strike}
  \vspace{-2.2em}
\end{figure*}

\margingraphics{figs/transformation.png}
\dbvisparagraph{Cluster Convex Hulls in 2D and HD} The cluster outputs for the 2D and HD data can be displayed as convex hulls \changed{i.e., the smallest convex polygon for which each point in the cluster is either on the boundary of the polygon or in its interior} \cite{sarikaya2018scatterplots}. Displaying HD clusters in the 2D projection can help capture the uncertainty that was introduced in the data transformation step. At the same time, if the HD clusters align with the 2D clusters, there is a higher likelihood that the clusters found in the projection also occur in the original HD data. 
\changed{\textit{Remark}: The resulting convex hulls of clusters in the HD space can have a loss of information, whereby the outliers can influence the hulls to cover a large area of the projection. However, these large areas can be a good visual hint to users, urging them to explore the particular neighborhoods.
}

\dbvisparagraph{Distance Matrices in 2D and HD} The second visual component for uncertainty communication of the data transformation is distance matrices.
The distance matrices of HD data can hint at projection issues. In particular, users can easily detect clusters visible in the matrix but not separated in the projection (see the distance matrix in HD for the third layer in \autoref{fig:distance-matrices}).
Inspired by the work of Cutura et al. \cite{cutura2020comparing}, we display pairwise distances between data points as rows and columns in a matrix. 
We use the \textit{viridis} color palette to encode the distance from dark blue representing small distances and yellow representing dissimilar data points.
To detect clusters, we support multiple matrix reordering techniques \cite{behrisch2016matrix} such as different linkage strategies, nearest-neighbor heuristics, and greedy heuristics. 
\changed{For linkage methods, hierarchical clustering is applied to the HD vectors, with the silhouette score used to identify the optimal cluster cut. The resulting dendrogram determines the ordering of data points, serving as the arrangement for the rows and columns in the matrix.}
\changed{The pseudocode of the reordering methods can be found in the supplementary material.} 
\changed{In addition, the bars on the top of the matrix represent the colors of the clusters in the top projection, and the left bars represent the colors of the clusters in the bottom projection.}

\begin{figure*}[t]
  \centering
  \includegraphics[width=\textwidth]{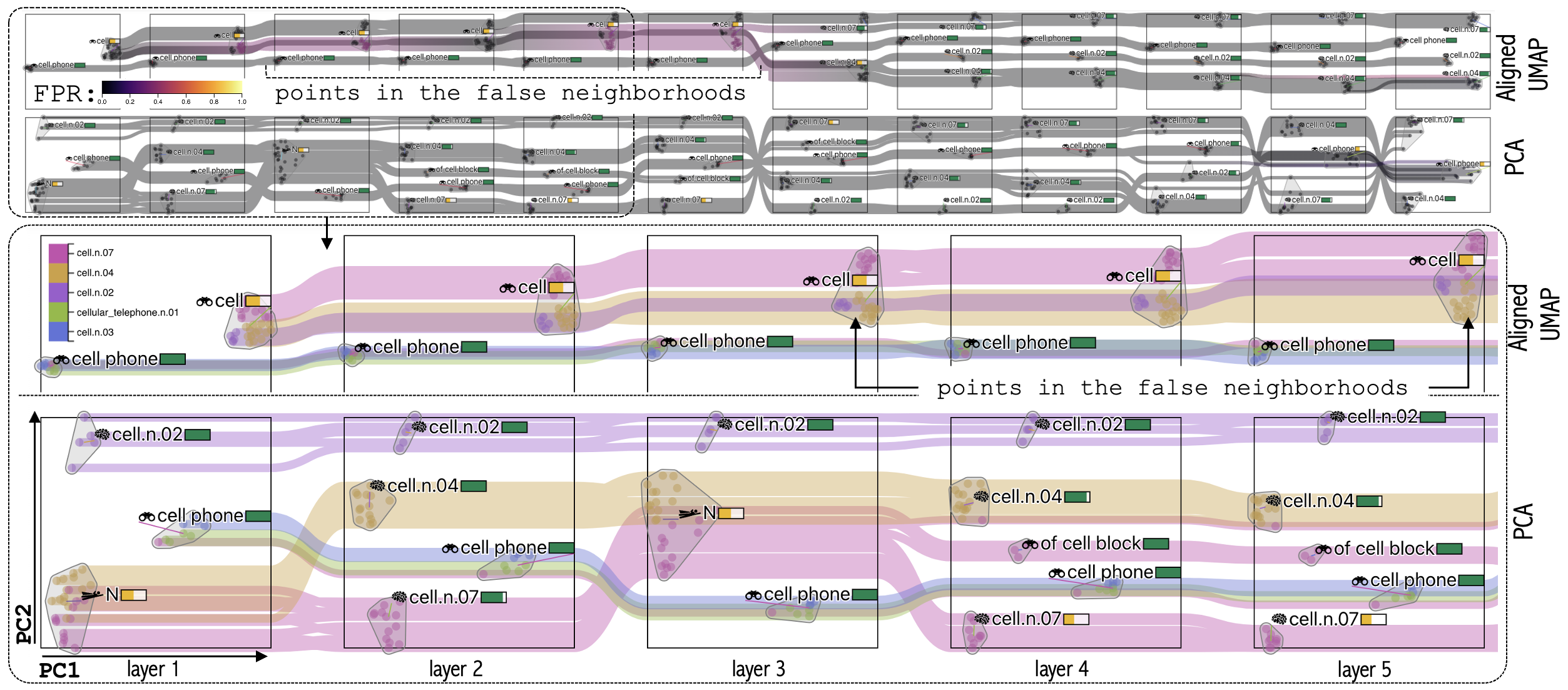}
  \vspace{-18pt}

  \caption{%
     The output of Aligned UMAP and PCA applied on \texttt{cell} tokens. The color of the zoomed-in flows shows the different word senses (e.g., \texttt{cell phone (n.03)}, \texttt{cellular telephone (n.01)}, \texttt{biological cell} (n.02), \texttt{prison cell (n.07)}, and \texttt{unit of political movement (n.04)}). The output of the PCA has a higher projection quality (see the dark grey color in the upper flows). In PCA, the first principal component (PC1) separates the \texttt{cell phones} from other word senses; the second principal component (PC2) in layers two, four, and five clearly separates \texttt{biological cells} from \texttt{prison cells}, whereby in layers four and five, \texttt{prison cells} that are mentioned in the context of ``cell block'' get further separated from the rest. 
  }
  \label{fig:use-case-cell}
  \vspace{-2.3em}
\end{figure*}

\margingraphics{figs/representation.png}
\dbvisparagraph{Flow Coloring according to Quality Metrics}
To communicate the uncertainty in the scatterplot representation itself, we use quality metrics that summarize the neighborhood artifacts that are potentially introduced by the dimensionality reduction method. These quality scores are represented for each data point using color encoding based on the \textit{interpolateInferno} color scale. The colored data points and the flows between layers emphasize regions with low quality, indicating that they should be interpreted cautiously, with careful consideration of the underlying data properties (see the violet flows in \autoref{fig:use-case-cell} for FPR).

\dbvisparagraph{K-Nearest Neighbor Connections} To validate the quality of each scatterplot, we connect k-nearest neighbor data points in the HD space through lines in the projection (shown in \autoref{fig:strike} \textbf{A}). This offers a visual cue on whether data point nearest neighbors are located in their surrounding 2D clusters. The parameter \textit{k} is specified by the user in the settings sidebar.

\margingraphics{figs/interpretation.png}
\dbvisparagraph{Cluster Summary Labels in 2D and HD}
Interpretation uncertainty can be reduced by providing users some hints about potential properties encoded in word embedding vectors \changed{(e.g., a common POS tag; see \autoref{sec:data-modeling}).} We compute the common property for data points in a cluster, display its label along the label's certainty value. The certainty is shown as a horizontal bar, whereby the color (green, yellow, red) maps the certainty value. 

\subsection{Close-Reading View}
Close reading is a fundamental method in text analysis applications~\cite{Janicke2015OnChallenges} explaining the reasons for the produced patterns in the analyzed data.
As shown in \autoref{fig:workspace}, the filtered data points can be explored in detail in a close-reading view.
The data points are grouped based on the detected 2D clusters annotated with the summary label and the word contexts. 

\section{Evaluation}
In the following, we showcase how the \textbf{LayerFlow} workspace can be used to explore embedding properties and how the visual components for uncertainty communication can highlight problems in the data transformation steps, distortions in the visual representations, and help to get insights during the interpretation phase.

\subsection{Study Setup}
The following insights were created through collaborative efforts with an expert in computational linguistics. The study was conducted via an online video conference; the expert was granted remote control access to the workspace. Initially, we assessed the expert's prior experience in working with language models and tasks related to word embeddings. The expert was then introduced to the workspace and its functionalities during a 10-minute briefing. Subsequently, the expert was asked to explore the contextualization of embeddings for the tokens \texttt{strike}, \texttt{cell}, and \texttt{post}. 
The expert was encouraged to talk aloud,  and their observations were recorded and used to create the following case studies.

\noindent{\textbf{Data} --} We use the Groningen Meaning Bank~\cite{Bos2017} dataset. This dataset provides a large collection of semantically annotated English texts with deep semantic representations. We use this corpus since it consists of many polysemous words used in diverse contexts, interesting for our particular use case.
\changed{The dataset contains 795,935 tokens (31,175 unique tokens) and 38,405 sentences.}

\noindent{\textbf{Model --}} In this study, we examine the embeddings of BERT language model, since it is one of the most widely researched encoder-based models and its embeddings are still commonly used for diverse text similarity tasks. It allows us to assess the viability of our workspace in replicating findings from previous studies concerning the language properties encoded within word embeddings across various layers of the model's architecture.

\subsection{Expert Case Studies}
Below, we present three case studies illustrating how the \textbf{LayerFlow} workspace facilitates gaining insights into embedding properties while also highlighting uncertainty in the resulting projections.

\noindent{\textbf{Aligned UMAP and PCA for \textit{strike} --}} \autoref{fig:strike} shows projections of the two middle layers of BERT created by the Aligned UMAP and PCA dimensionality reduction methods applied to a set of words referred to as \texttt{strike}. First, by exploring the k-nearest neighbors (k=1), the expert noticed that PCA produces a 2D space that poorly captures the HD neighborhoods (see \autoref{fig:strike} \textbf{A}, the two lower projections). Many lines were connecting data points of two separate clusters. Moreover, the close neighborhoods within one cluster were also poorly represented as shown by the long lines between data points within a convex hull. 
By connecting words through the flow diagram (see \autoref{fig:strike} \textbf{B} and \textbf{C}, the two lower projections), the issue of separating clusters became even more apparent. 
Different to PCA, the projections produced by the Aligned UMAP presented three distinct clusters (see the upper projections in \autoref{fig:strike}). Data points that were split into multiple groups tended to belong to these groups also in the HD space. 

\begin{figure*}[t]
  \centering
  \includegraphics[width=\textwidth]{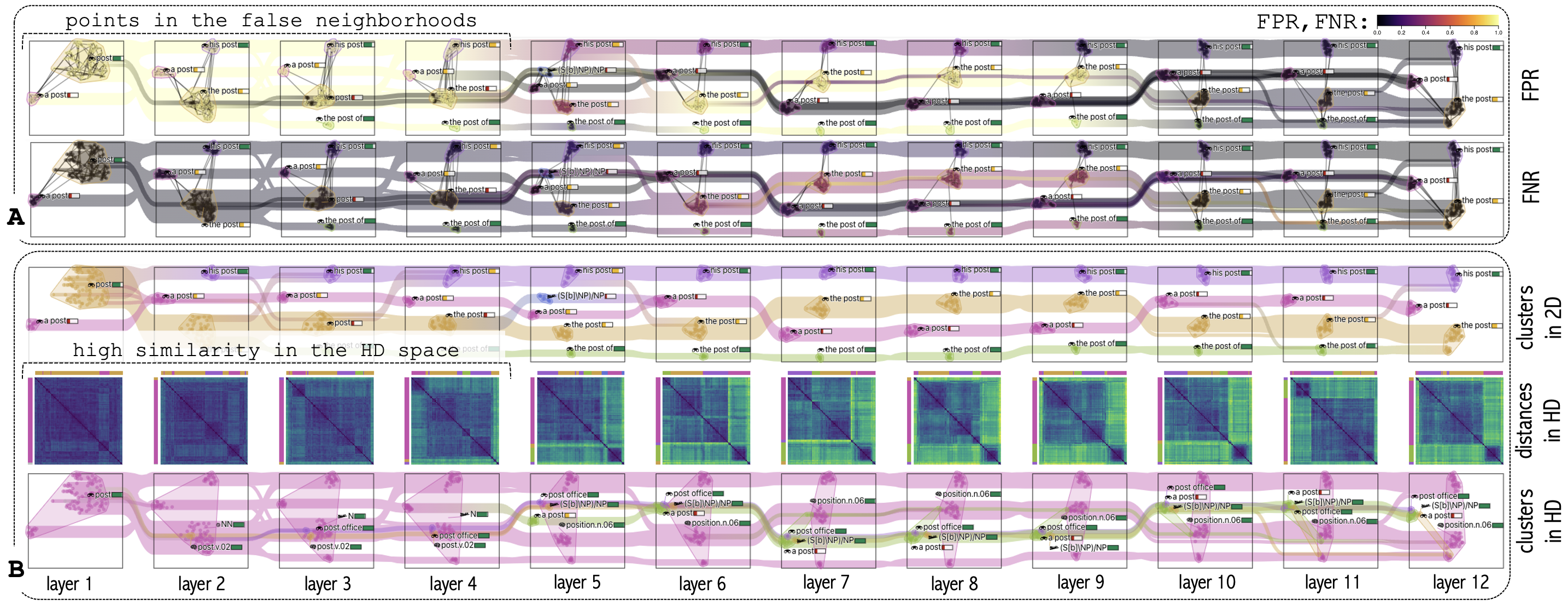}
  \vspace{-18pt}

  \caption{%
    As shown in \textbf{A} through the bright colors (i.e., high FPR and FNR values), Aligned UMAP has challenges representing the HD neighborhoods, i.e., many points occur in the false neighborhoods. The convex-hulls show that there is a misalignment between the clusters in the 2D (shown in \textbf{B}, above the matrices) and HD space (\textbf{B}, below the matrices). The connections between the k-nearest neighbors (in \textbf{A}) and distance matrices (in \textbf{B}) reveal that data points in early layers (i.e., 1-4) are very similar in the HD space. The Aligned UMAP, however, captures minor embedding differences and produces more distinct clusters than the hierarchical clustering on the HD vectors. 
  }
  \label{fig:use-case-post}
  \vspace{-2.3em}
\end{figure*}

\noindent{\textbf{Aligned UMAP and PCA for \textit{cell} --}}
\autoref{fig:use-case-cell} displays the projections of Aligned UMAP and PCA applied to a set of words \texttt{cell}. The expert aimed to explore whether embeddings encode the word meaning. Therefore, they first colored the flows according to the different word senses (e.g., \texttt{cell phone (n.03)}, \texttt{cellular telephone (n.01)}, \texttt{biological cell (n.02)}, \texttt{prison cell (n.07)}, and \texttt{unit of political movement (n.04)}). The expert noticed that in the first layers of BERT, the Aligned UMAP separated the \texttt{cell phones} and \texttt{cellular telephone} from other word senses and placed the three word senses (i.e., \texttt{n.07}, \texttt{n.04}, \texttt{n.02}) in the neighborhood, which, due to the small distances between the data points, were grouped together into one cluster by the clustering algorithm. As noticed by the expert, this reduced the projection quality scores, as shown by the purple color in the center of \autoref{fig:use-case-cell}. However, the expert observed that despite the close proximity of the data points, the dimensionality reduction algorithm managed to separate the three word senses. Thus, the quality issues highlighted by the quality metric was an artifact of the applied clustering algorithm. To verify this observation, the expert additionally displayed the projections of the PCA algorithm.
In PCA, the first principal component separated the \texttt{cell phones} from other word senses (see the x-axis in the visualization). This was similar to the output of the Aligned UMAP. The second principal component (see the y-axis) in layers two, four, and five, however, separated \texttt{biological cells} from \texttt{prison cells}, whereby in layers four and five, \texttt{prison cells} that were mentioned in the context of \texttt{cell block} were further separated from the rest.
With some minor exceptions, the FPR score had low values for most of the 2D clusters. By observing further quality metrics, the expert concluded that both projections produced outputs that were relatively comparable and trustworthy.
The main difference was that PCA tended to separate word senses in even smaller clusters having common n-grams in their word contexts (e.g., ``of cell block''). Apparently, such groups of common n-grams can be detected through a linear combination of the embedding dimensions.

\noindent{\textbf{Aligned UMAP for \textit{post} --}} \texttt{Post} is a polysemous word that has multiple possible meanings such as the \texttt{institution}, \texttt{position in a company}, or \texttt{the action of publishing}. \autoref{fig:use-case-post} shows the projections created by the Aligned UMAP algorithm on a set of these polysemous words. The expert started by inspecting the different quality metric outputs. As shown in \autoref{fig:use-case-post} \textbf{A}, the FPR and FNR quality scores highlighted issues related to the projection neighborhood preservation. In the first nine layers, many data points were assigned to neighborhoods that did not exist in the HD space; in the middle layers (six to nine), some nearest neighbors that existed in the HD space were not recognized in the 2D space. The expert aimed to understand the reasons. Therefore, they displayed the Aligned UMAP outputs twice and compared the clusters in both 2D and HD space. In addition, the expert displayed the distance matrices in the HD space to be able to interpret the original HD data and potential data transformation uncertainties.

As shown in \autoref{fig:use-case-post} \textbf{B}, most of the data points in the early layers had a high similarity. Thus, the clustering algorithm on the HD vectors grouped all data points into one cluster except a few outliers. 
The distance matrices showed that the Aligned UMAP separated clusters based on smaller data differences; thus, in many layers, there was a difference between the number of found 2D and HD clusters. The expert concluded that the clustering results must have influenced the quality metrics that highlighted projection quality issues. To understand whether this was an artifact of the HD clustering approach and the projections produced meaningful results, they compared the groups visible in the distance matrix to the 2D clusters in the projections. Although some correlations could be detected, i.e., some 2D clusters could be recognized in the ordering of the distance matrices, the distances were not large enough to make strong conclusions. Thus, the expert decided to verify this further based on the k-nearest neighborhoods and displayed the k-nearest neighbors in the HD space. As shown in \autoref{fig:use-case-post} \textbf{A}, connecting lines of the k-nearest neighbors confirmed the quality issues; many data points were assigned to clusters different to their nearest neighbors in the HD space. The expert concluded that, apparently, multiple visual hints are needed to assess the projection quality issues and understand the potential reasons thereof.

\begin{figure*}[t]
  \centering
  \includegraphics[width=\linewidth]{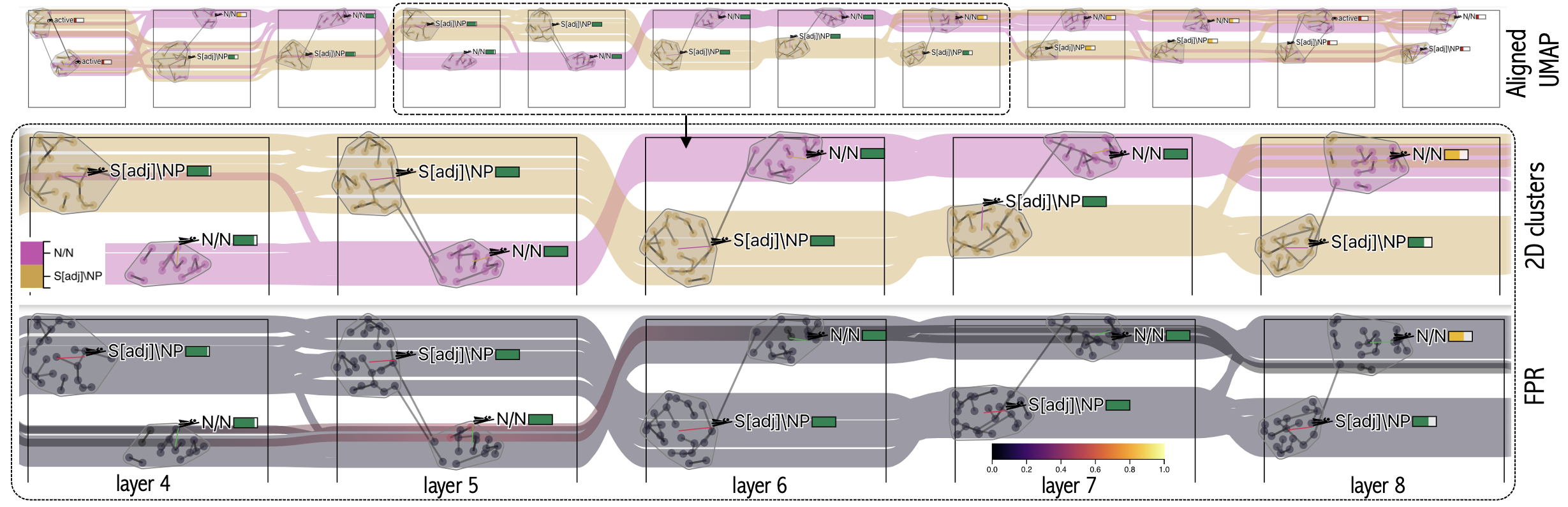}
  \vspace{-20pt}
  \caption{%
    Aligned UMAP for the word \texttt{active}. The visualization shows that especially in the middle layers of BERT, words are grouped by their syntactic categories (here, \texttt{N/N} and \texttt{S[adj]\textbackslash NP}), conforming findings from prior work \cite{rogers2020primer,sevastjanova2022lmfingerprints}. There are only 1-2 words that are clustered in a different group than their 1st nearest neighbor in the HD space. The projection quality is further confirmed by the quality metric (here, the FPR shows low values, i.e., the data point neighborhoods have a small False Positive Rate). 
  }
  \label{fig:use-case-syntax}
  \vspace{-2.1em}
\end{figure*}

\subsection{Replicating Findings from Related Work}
The BERTology paper~\cite{rogers2020primer} and LMFingerprints~\cite{sevastjanova2022lmfingerprints} show that BERT encodes POS tag information in the early to middle layers (2-5) and more complex syntactic structures in the middle layers of its architecture.
To verify whether our workspace shows similar patterns, we select the polysemous word \texttt{active}.

\texttt{Active} can be used as an adjective in two different ways. Adjectives are usually placed before the nouns they modify (e.g., \texttt{It is the northernmost active volcano (N) on earth.}), but when used with linking verbs, they are placed after the verb (e.g., \texttt{The hidden volcano is (VB) still active.}). The latter type of adjective is called a predicative adjective. \autoref{fig:use-case-syntax} shows the output of the Aligned UMAP for a sample of words \texttt{active} used in different contexts. As shown by the two distinct colors, the middle layers group these words based on their syntactic category. In particular, words that are used before the nouns they modify are grouped together (see \texttt{N/N}), and predicative adjectives are grouped in a separate cluster (see \texttt{S[adj]\textbackslash NP}). We display the connecting lines between the k-nearest neighbors and the FPR quality metric to highlight that the produced 2D clusters indeed match the clusters in the HD space.

\subsection{Preliminary Expert Feedback}
Besides the computational linguistics expert, we introduced the tool to three visual analytics experts for preliminary feedback on its visual and interaction design. 
In the following, we summarize the main feedback gathered during semi-structured interviews.

The experts acknowledged the visual design of the interlinked projections. Asked to compare the different layout alternatives (the three steps described in \autoref{sec:interlinked-projections}, i.e., simple lines vs. Sankey diagram), all experts recognized the Sankey layout as the most effective approach to gain insights into cluster and neighborhood changes. 
The experts expressed the benefit of combining multiple visual components and struggled naming one visual representation that would outperform others. The expert in computational linguists suggested starting with a combination of all visual components rather with a blank canvas (the setting used during the expert studies) and requiring the users to deselect components that they find less relevant for the particular data used.
Overall, the experts expressed positive feedback regarding the functionality and the visual design of the workspace. The possibility to select words, display embeddings layer-wise, get an overview of trends, projection quality, and summary annotations received a positive feedback. 
Also, several suggestions for improvements and future work were provided such as integrating further linguistic features into summary visualizations and providing more accessible reasons for projection quality issues (i.e., the explanation of the FPR and FNR computation), which should be addressed in the future work.

\section{Discussion and Future Work}

\changed{
With this work, we aim to motivate the users to exercise caution when analyzing word embeddings and drawing conclusions about the model's learning capabilities from visual patterns that may not truly reflect the underlying data distribution. 
Due to the different types of distortions and uncertainties possible, we keep the users in charge of exploring the visual representations and concluding whether the insights are trustworthy.
As described in \autoref{sec:design-rationale}, when a projection produces high uncertainty, the users can select another projection technique or adapt its parameters to explore whether the results are more trustworthy. Our case studies also show that in some situations, it might be meaningful to use multiple projection methods simultaneously, e.g., PCA on the embedding extracted from early layers and a non-linear projection technique on the upper layers' embeddings.
}
In the following, we describe some opportunities for future work.

\noindent{\textbf{Automatic Suggestion of Uncertainty Components --}} Currently, the user has to decide which visual components to display for uncertainty communication. This requires the user to decide when which components are necessary. Future work should explore the possibilities to automatically suggest relevant components based on the underlying data properties. E.g., if the data points are difficult to separate in the HD space, then the system should automatically warn the user of potential misrepresentation. Guidance in visual analytics systems \cite{sperrle2020} becomes more important and should be used also to communicate visual uncertainties.

\noindent{\textbf{Quality Metrics --}} The interpretation of the quality metric results was mentioned as one of the most challenging aspects of the workspace. Although the experts mentioned that they are helpful to assess which projections represent the HD data better than others, they would prefer to see direct impact of the different nearest neighbors for the metric values. Although the k-nearest neighbor connections were judged useful, they do not explain the quality metric values. Since the FPR, FNR metrics are computed based on the minimum spanning tree of the LD space, a visual explanation related to the spanning tree would ease their interpretation.

\noindent{\textbf{Scalability --}} The presented examples were limited to embedding subsets ranging to 150 data points per layer. Such data selection is common for word embedding analysis approaches. Nevertheless, the more data points are used, the more space is required for the distance matrix visualization, thus limiting the information that can be displayed on the screen simultaneously. Moreover, the more data points are displayed, the more flows need to be visualized, potentially reducing their readability. 
\changed{Future research should explore how to scale the visual representation to more data points as well as more layers. For instance, one could create multiple abstraction levels of the displayed information and uncertainty and use a semantic zoom to navigate through the visual space.}

\noindent{\textbf{Quality of HD Clusters --}} HD clusters and neighborhoods in the HD space are used as basis to compute the projection quality metrics (see \autoref{sec:data-modeling}). The quality of these metrics, thus, depends on the quality of the clustering output. If the clustering has difficulties in separating embeddings into distinct clusters, it will become apparent in the quality scores. If the problem is the data itself, then this is an expected result, i.e., the user should be alerted that there are some difficulties in separating the embeddings into groups. However, if the problem is the clustering algorithm (and there are distinct groups in the HD space), then this might restrict the user in gaining fast insights about the data properties.

\section{Conclusion}
In this paper, we present a workspace called \textbf{LayerFlow} that visualizes word embedding vectors layer-wise and supports analysis in exploring their encoded linguistic properties. In this work, we build on our previous work~\cite{sevastjanova2021} that introduced interlinked projections for embedding analysis, present a new flow-design that supports the investigation of cluster changes between consecutive layers, and integrate several visual components to communicate the uncertainty. In particular, we address the uncertainty that can occur within the data transformation, representation, and interpretation steps of the analysis. A demo is available under \href{https://layerflow.ivia.ch}{layerflow.ivia.ch}.

\section*{Acknowledgments}
\noindent This paper was funded by the Swiss National Science Foundation (SNSF) within the project 10003068.  

\clearpage
\printbibliography
\end{document}